# Gate-Variants of Gated Recurrent Unit (GRU) Neural Networks

Rahul Dey and Fathi M. Salem

*Circuits, Systems, and Neural Networks (CSANN) LAB*
*Department of Electrical and Computer Engineering*
*Michigan State University*
*East Lansing, MI 48824-1226, USA*

deyrahul@msu.edu || salemf@msu.edu

*Abstract* – **The paper evaluates three variants of the Gated Recurrent Unit (GRU) in recurrent neural networks (RNN) by reducing parameters in the update and reset gates. We evaluate the three variant GRU models on MNIST and IMDB datasets and show that these GRU-RNN variant models perform as well as the original GRU RNN model while reducing the computational expense.**

## I. INTRODUCTION

Gated Recurrent Neural Network (RNN) have shown success in several applications involving sequential or temporal data [1-13]. For example, they have been applied extensively in speech recognition, natural language processing, machine translation, etc. [2, 5]. Long Short-Term Memory (LSTM) RNN and the recently introduced Gated Recurrent Unit (GRU) RNN have been successfully shown to perform well with long sequence applications [2-5, 8-12].

Their success is primarily due to the gating network signals that control how the present input and previous memory are used to update the current activation and produce the current state. These gates have their own sets of weights that are adaptively updated in the learning phase (i.e., the training and evaluation process). While these models empower successful learning in RNN, they introduce an increase in parameterization through their gate networks. Consequently, there is an added computational expense vis-à-vis the simple RNN model [2, 5, 6]. It is noted that the LSTM RNN employs 3 distinct gate networks while the GRU RNN reduce the gate networks to two. In [14], it is proposed to reduce the external gates to the minimum of one with preliminary evaluation of sustained performance.

In this paper, we focus on the GRU RNN and explore three new gate-variants with reduced parameterization. We comparatively evaluate the performance of the original and the variant GRU RNN on two public datasets. Using the MNIST dataset, one generates two sequences [2, 5, 6, 14]. One sequence is obtained from each 28x28 image sample as pixel-wise long sequence of length 28x28=784 (basically, scanning from the upper left to the bottom right of the image). Also, one generates a row-wise short sequence of length 28, with each element being a vector of dimension 28 [14, 15]. The third sequence type employs the IMDB movie review dataset where one defines the length of the sequence in order to achieve high performance sentiment classification from a given review paragraph.

## II. BACKGROUND: RNN, LSTM AND GRU

In principal, RNN are more suitable for capturing relationships among sequential data types. The so-called simple RNN has a recurrent hidden state as in

$$h_t = g(Wx_t + Uh_{t-1} + b) \qquad (1)$$

where $x_t$ is the (external) *m-dimensional* input vector at time $t$, $h_t$ the *n-dimensional* hidden state, g is the (point-wise) activation function, such as the logistic function, the hyperbolic tangent function, or the rectified Linear Unit (ReLU) [2, 6], and $W, U$ and $b$ are the appropriately sized parameters (two weights and bias). Specifically, in this case, $W$ is an $n \times m$ matrix, $U$ is an $n \times n$ matrix, and $b$ is an $n \times 1$ matrix (or vector).

Bengio at al. [1] showed that it is difficult to capture long-term dependencies using such simple RNN because the (stochastic) gradients tend to either vanish or explode with long sequences. Two particular models, the Long Short-Term Memory (LSTM) unit RNN [3, 4] and Gated Recurrent Unit (GRU) RNN [2] have been proposed to solve the "vanishing" or "exploding" gradient problems. We will present these two models in sufficient details for our purposes below.

### A. Long Short-Term Memory (LSTM) RNN
The LSTM RNN architecture uses the computation of the simple RNN of Eqn (1) as an intermediate candidate for the internal memory cell (state), say $\tilde{c}_t$, and add it in a (element-wise) weighted-sum to the previous value of the internal memory state, say $c_{t-1}$, to produce the current value of the memory cell (state) $c_t$. This is expressed succinctly in the following discrete dynamic equations:

$$c_t = f_t \odot c_{t-1} + i_t \odot \tilde{c}_t \qquad (2)$$
$$\tilde{c}_t = g(W_c x_t + U_c h_{t-1} + b_c) \qquad (3)$$
$$h_t = o_t \odot g(c_t) \qquad (4)$$

In Eqns (3) and (4), the activation nonlinearity $g$ is typically the hyperbolic tangent function but more recently may be implemented as a rectified Linear Unit (reLU). The weighted sum is implemented in Eqn (2) via element-wise (Hadamard) multiplication denoted by ⊙ to gating signals. The gating (control) signals $i_t$, $f_t$ and $o_t$ denote, respectively, the *input*, *forget*, and *output* gating signals at time $t$. These control gating signals are in fact replica of the basic equation (3), with their own parameters and replacing $g$ by the logistic function. The logistic function limits the gating signals to within 0 and 1. The specific mathematical form of the gating signals are thus expressed as the vector equations:

$$i_t = \sigma(W_i x_t + U_i h_{t-1} + b_i)$$
$$f_t = \sigma(W_f x_t + U_f h_{t-1} + b_f)$$
$$o_t = \sigma(W_o x_t + U_o h_{t-1} + b_o)$$

where $\sigma$ is the logistic nonlinearity and the parameters for each gate consist of two matrices and a bias vector. Thus, the total number of parameters (represented as matrices and bias vectors) for the 3 gates and the memory cell structure are, respectively, $W_i, U_i, b_i, W_f, U_f, b_f, W_o, U_o, b_o, W_c, U_c$ and $b_c$. These parameters are all updated at each training step and stored. It is immediately noted that the number of parameters in the LSTM model is increased 4-folds from the simple RNN model in Eqn (1). Assume that the cell state is *n-dimensional*. (Note that the activation and all the gates have the same dimensions). Assume also that the input signal is *m-dimensional*. Then, the total parameters in the LSTM RNN are equal to $4\times(n^2 + nm + n)$.

*B. Gated Recurrent Unit (GRU) RNN*

The GRU RNN reduce the gating signals to two from the LSTM RNN model. The two gates are called an update gate $z_t$ and a reset gate $r_t$. The GRU RNN model is presented in the form:

$$h_t = (1-z_t) \odot h_{t-1} + z_t \odot \tilde{h}_t \quad (5)$$
$$\tilde{h}_t = g(W_h x_t + U_h(r_t \odot h_{t-1}) + b_h) \quad (6)$$

with the two gates presented as:
$$z_t = \sigma(W_z x_t + U_z h_{t-1} + b_z) \quad (7)$$
$$r_t = \sigma(W_r x_t + U_r h_{t-1} + b_r) \quad (8)$$

One observes that the GRU RNN [Eqns (5)-(6)] is similar to the LSTM RNN [Eqns (2)-(3)], however with less external gating signal in the interpolation Eqn (5). This saves one gating signal and the associated parameters. We defer further information to reference [2], and the references therein. In essence, the GRU RNN has 3-folds increase in parameters in comparison to the simple RNN of Eqn (1). Specifically, the total number of parameters in the GRU RNN equals $3\times(n^2 + nm + n)$.

In various studies, e.g., in [2] and the references therein, it has been noted that GRU RNN is comparable to, or even outperforms, the LSTM in most cases. Moreover, there are other reduced gated RNNs, e.g. the Minimal Gated Unit (MGU) RNN, where only one gate equation is used and it is reported that this (MGU) RNN performance is comparable to the LSTM RNN and the GRU RNN, see [14] for details.

In this paper, we focus on the GRU RNN model and evaluate new variants. Specifically, we retain the architecture of Eqns (5)-(6) unchanged, and focus on variation in the structure of the gating signals in Eqns (7) and (8). We apply the variations identically to the two gates for uniformity and simplicity.

### III. THE VARIANT GRU ARCHITECTURES

The gating mechanism in the GRU (and LSTM) RNN is a replica of the simple RNN in terms of parametrization. The weights corresponding to these gates are also updated using the backpropagation through time (BTT) stochastic gradient descent as it seeks to minimize a loss/cost function [3, 4]. Thus, each parameter update will involve information pertaining to the state of the overall network. Thus, all information regarding the current input and the previous hidden states are reflected in the latest state variable. There is a redundancy in the signals driving the gating signals. The key driving signal should be the internal state of the network. Moreover, the adaptive parameter updates all involve components of the internal state of the system [16, 17]. In this study, we consider three distinct variants of the gating equations applied uniformly to both gates:

**Variant 1**: called **GRU1**, where each gate is computed using only the previous hidden state and the bias.

$$z_t = \sigma(U_z h_{t-1} + b_z) \quad (9-a)$$
$$r_t = \sigma(U_r h_{t-1} + b_r) \quad (9-b)$$

Thus, the total number of parameters is now reduced in comparison to the GRU RNN by $2\times nm$.

**Variant 2:** called **GRU2**, where each gate is computed using only the previous hidden state.
$$z_t = \sigma(U_z h_{t-1}) \quad (10-a)$$
$$r_t = \sigma(U_r h_{t-1}) \quad (10-b)$$

Thus, the total number of parameters is reduced in comparison to the GRU RNN by $2\times (nm+n)$.

**Variant 3:** called **GRU3**, where each gate is computed using only the bias.
$$z_t = \sigma(b_z) \quad (11-a)$$
$$r_t = \sigma(b_r) \quad (11-b)$$

Thus the total number of parameters is reduced in comparison to the GRU RNN by $2\times (nm+n^2)$.

We have performed an empirical study of the performance of each of these variants as compared to the GRU RNN on, first, sequences generated from the MNIST dataset and then on the IMDB movie review dataset. In the subsequent figures and tables, we refer to the base GRU RNN model as GRU0 and the three variants as GRU1, GRU2, and GRU3 respectively.

Our architecture consists of a single layer of one of the variants of GRU units driven by the input sequence and the activation function $g$ set as ReLU. (Initial experiments using

$g = tanh$ have produced similar results). For the MNIST dataset, we generate the pixel-wise and the row-wise sequences as in [15]. The networks have been generated in Python using the Keras library [15] with Theano as a backend library. As Keras has a GRU layer class, we modified this class to classes for GRU1, GRU2, and GRU3. All of these classes used the ReLU activation function. The RNN layer of units is followed by a softmax layer in the case of the MNIST dataset or a traditional logistic activation layer in the case of the IMDB dataset to predict the output category. The Root Mean Square Propagation (RMSprop) is used as the choice of optimizer that is known to adapt the learning rate for each of the parameters. To speed up training, we also decay the learning rate exponentially with the cost in each epoch

$$\eta = \eta_0 e^{cost} \qquad (12)$$

where $\eta_0$ represents a base constant learning rate and $cost$ is the cost computed in the previous epoch. The details of our models are delineated in Table I.

Table I: Network model characteristics

| Model | MNIST Pixel-wise | MNIST Row-wise | IMDB |
|---|---|---|---|
| Hidden Units | 100 | 100 | 128 |
| Gate Activation | Sigmoid | Sigmoid | sigmoid |
| Activation | ReLU | ReLU | ReLU |
| Cost | Categorical Cross-entropy | Categorical Cross-entropy | Binary Cross-entropy |
| Epochs | 100 | 50 | 100 |
| Optimizer | RMSProp | RMSProp | RMSProp |
| Dropout | 20% | 20% | 20% |
| Batch Size | 32 | 32 | 32 |

IV. RESULTS AND DISCUSSION

A. Application to MNIST Dataset – pixel-wise sequences

The MNIST dataset [15] consists of 60000 training images and 10000 test images, each of size 28x28 of handwritten digits. We evaluated our three variants against the original GRU model on the MNIST dataset by generating the sequential input in one case (pixel-wise, one pixel at a time) and in the second case (row-wise, one row at a time). The pixel-wise sequence generated from each image are 1-element signal of length 784, while the 28-element row-wise produces a sequence of length 28. For each case, we performed different iterations by varying the constant base learning rate $\eta_0$. The results of our experiments are depicted in Fig. 1, 2, and 3, with summary in Table II below.

Table II: MNIST pixel-wise sequences: performance summary of different architectures using 4 constant base learning rates $\eta_0$, in 100 epochs.

| Architecture | Lr = 1e-3 Train | Lr = 1e-3 Test | Lr = 5e-4 Train | Lr = 5e-4 Test | 1e-4 Train | 5e-5 Test | # Params |
|---|---|---|---|---|---|---|---|
| GRU0 | 99.19 | 98.59 | 98.59 | 98.04 | - | - | 30600 |
| GRU1 | 98.88 | 98.37 | 98.52 | - | - | - | 30400 |
| GRU2 | 98.90 | 98.10 | 98.61 | - | - | - | 30200 |
| GRU3 | - | - | 10.44 | - | 60.97 | 59.60 | 10400 |

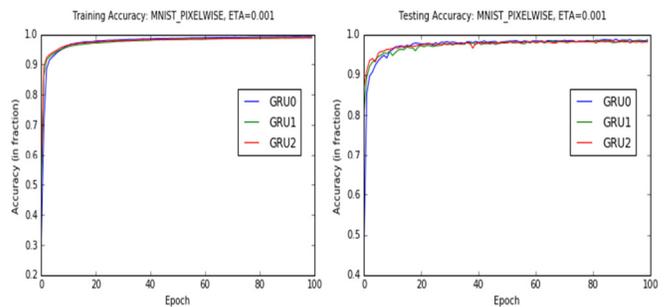

Fig. 1 Training (left) and Testing (right) Accuracy of GRU0, GRU1 and GRU2 on MNIST pixel-wise generated sequences at eta=0.001

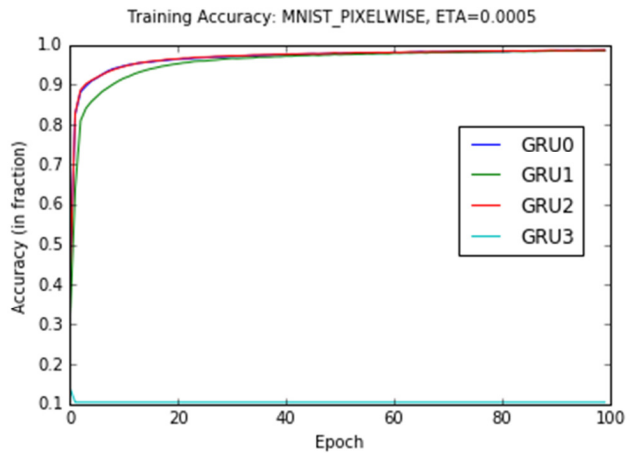

Fig. 2 Training Accuracy of GRU0, GRU1, GRU2 and GRU3 on MNIST generated sequences at eta=5e-4

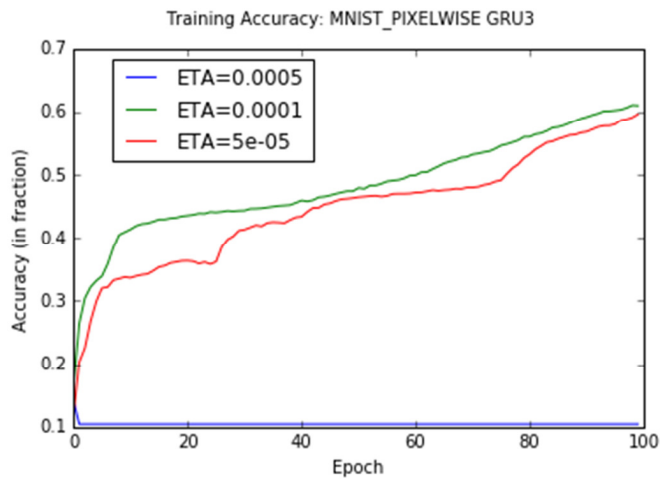

Fig. 3 Performance of GRU3 on MNIST generated sequences for 3 constant base learning rates $\eta_0$.

From Table II and Fig. 1 and 2, GRU1 and GRU2 perform almost as well as GRU0 on MNIST pixel-wise generated sequence inputs. While GRU3 does not perform as well for this (constant base) learning rate. Figure 3 shows that reducing the (constant base) learning rate to (0.0001) and below has enabled GRU3 to increase its (test) accuracy performance to 59.6% after 100 epochs, and with a positive slope indicating that it would increase further after more epochs. Note that in this experiment, GRU3 has about 33% of the number of (adaptively computed) parameters compared to GRU0. Thus, there exists a potential trade-off between the higher accuracy performance and the decrease in the number of parameters. In our experiments, using 100 epochs, the GRU3 architecture never attains saturation. Further experiments using more epochs and/or more units would shed more light on the comparative evaluation of this trade-off between performance and parameter-reduction.

*B. Application to MNIST Dataset – row-wise sequences*

While pixel-wise sequences represent relatively long sequences, row-wise generated sequences can test short sequences (of length 28) with vector elements. The accuracy profile performance vs. epochs of the MNIST dataset with row-wise input of all four GRU RNN variants are depicted in Fig. 4, Fig. 5, and Fig. 6, using several constant base learning rates. Accuracy performance results are then summarized in Table III below.

Table III: MNIST row-wise generated sequences: Accuracy (%) performance of different variants using several constant base learning rates over 50 epochs

|  | Lr = 1e-2 | | Lr = 1e-3 | | Lr = 1e-4 | | # Params |
|---|---|---|---|---|---|---|---|
| Architecture | Train | Test | Train | Test | Train | Test | |
| GRU0 | 96.99 | 98.49 | 98.14 | 98.85 | 93.02 | 96.66 | 38700 |
| GRU1 | 97.24 | 98.55 | 97.46 | 98.93 | 91.54 | 96.58 | 33100 |
| GRU2 | 96.95 | 98.71 | 97.33 | 98.93 | 91.20 | 96.23 | 32900 |
| GRU3 | 97.19 | 98.85 | 97.04 | 97.39 | 80.33 | 87.96 | 13100 |

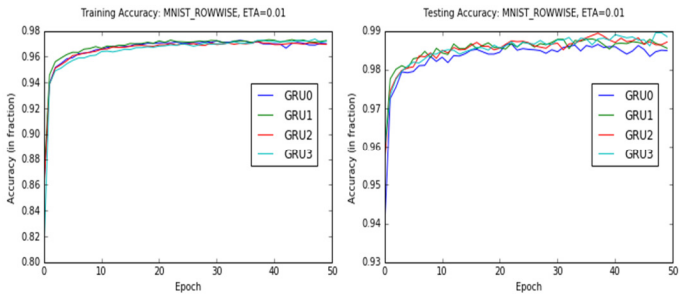

Fig. 4 Training and testing accuracy on MNIST row-wise generated sequences at a constant base learning rate of 1e-2

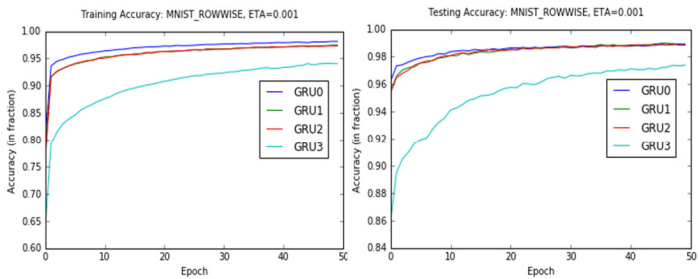

Fig. 5 Training and testing accuracy on MNIST row-wise generated sequences at a constant base learning rate of 1e-3

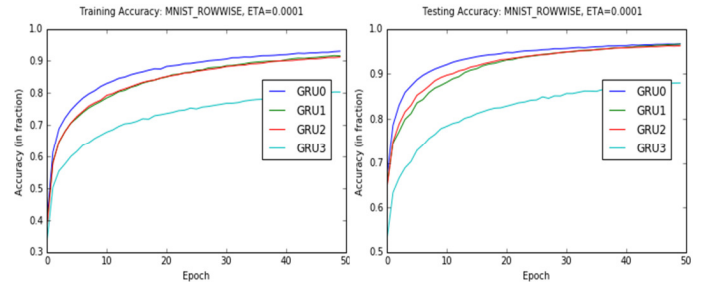

Fig. 6 Training and testing accuracy on MNIST row-wise generated sequences at a constant base learning rate of 1e-4

From Table III and Fig. 4, Fig.5, and Fig. 6, all the four variants GRU0, GRU1, GRU2, and GRU3 appear to exhibit comparable accuracy performance over three constant base learning rates. GRU3 exhibits lower performance at the base learning rate of 1e-4 where, after 50 epochs, is still lagging. From Fig. 6, however, it appears that the profile has not yet levelled off and has a positive slope. More epochs are likely to increase performance to comparable levels with the other variants. It is noted that in this experiment, GRU3 can achieve comparable performance with roughly one third of the number of (adaptively computed) parameters. Computational expense savings may play a role in favoring one variant over the others in targeted applications and/or available resources.

*C. Application to the IMDB Dataset– natural sequence*

The IMDB dataset is composed of 25000 test data and 25000 training data consisting of movie reviews and their binary sentiment classification. Each review is represented by a maximum of 80 (most frequently occurring) words in a vocabulary of 20000 words [7]. We have trained the dataset on all 4 GRU variants using the two constant base learning rates of 1e-3 and 1e-4 over 100 epochs. In the training, we employ *128-dimensional* GRU RNN variants and have adopted a batch size of 32. We have observed that, using the constant base learning rate of 1e-3, performance fluctuates visibly (see Fig. 7), whereas performance is uniformly progressing over profile-curves as shown in Fig. 8. Table IV summarizes the results of accuracy performance which show comparable performance among GRU0, GRU1, GRU2, and GRU3. Table IV also lists the number of parameters in each.

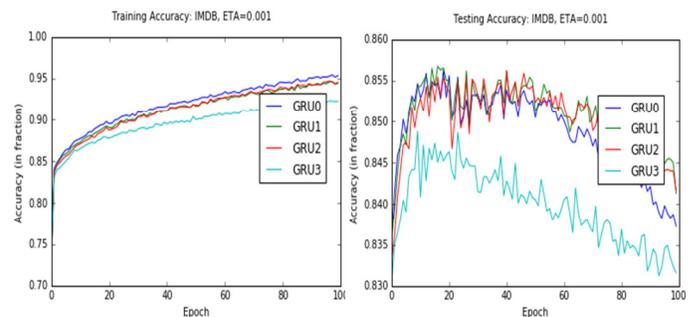

Fig. 7 Test and validation accuracy on IMDB dataset using a base learning rate of 1e-3

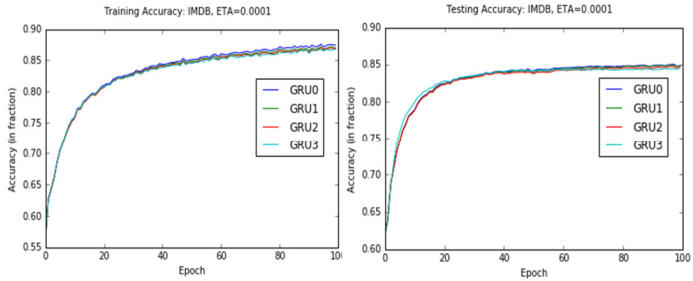

Fig. 8 Training and testing accuracy on IMDB dataset using a base learning rate of 1e-4

Table IV: IMDB dataset: Accuracy (%) performance of different architectures using two base learning rates over 100 epochs

| Architecture | Lr = 1e-3 Train | Lr = 1e-3 Test | Lr = 1e-4 Train | Lr = 1e-4 Test | # Params |
|---|---|---|---|---|---|
| GRU0 | 95.3 | 83.7 | 87.4 | 84.8 | 98688 |
| GRU1 | 94.5 | 84.1 | 87.0 | 84.8 | 65920 |
| GRU2 | 94.5 | 84.2 | 86.9 | 84.6 | 65664 |
| GRU3 | 92.3 | 83.2 | 86.8 | 84.5 | 33152 |

The IMDB data experiments provide the most striking results. It can be clearly seen that all the 3 GRU variants perform comparably to the GRU RNN while using less number of parameters. The learning pace of GRU3 was also similar to those of the other variants at the constant base learning rate of 1e-4. From Table IV, it is noted that more saving in computational load is achieved by all variant GRU RNN as the input is represented as a large 128-*dimensional* vector.

## V. CONCLUSION

The experiments on the variants GRU1, GRU2, and GRU3 verse the GRU RNN have demonstrated that their accuracy performance is comparable on three example sequence lengths. Two sequences generated from the MNIST dataset and one from the IMDB dataset. The main driving signal of the gates appear to be the (recurrent) state as it contains essential information about other signals. Moreover, the use of the stochastic gradient descent implicitly carries information about the network state. This may explain the relative success in using the bias alone in the gate signals as its adaptive update carries information about the state of the network. The GRU variants reduce this redundancy and thus their performance has been comparable to the original GRU RNN. While GRU1 and GRU2 have indistinguishable performance from the GRU RNN, GRU3 frequently lags in performance, especially for relatively long sequences and may require more execution time to achieve comparable performance,

We remark that the goal of this work to comparatively evaluate the performance of GRU1, GRU2 and GRU3, which possess less gate parameters, and thus less computational expense, than the original GRU RNN. By performing more experimental evaluations using constant or varying learning rates, and training for longer number of epochs, one can validate the performance on broader domain. We remark that the three GRU RNN variants need to be further comparatively evaluated on diverse datasets for a broader empirical performance evidence.